\documentclass[runningheads]{llncs}

\usepackage{bbding}
\usepackage{tabularx}
\usepackage{overpic}
\usepackage[dvipsnames]{xcolor}

\usepackage[mobile]{eccv}

\usepackage{eccvabbrv}

\usepackage{graphicx}
\usepackage{booktabs}

\usepackage[accsupp]{axessibility}  

\definecolor{darkred}{rgb}{0.7, 0.0, 0.0}
\definecolor{darkgreen}{rgb}{0.0, 0.37, 0.14}
\definecolor{darkblue}{rgb}{0.10, 0.17, 0.8}

\newcommand{\myparagraph}[1]{\noindent\textbf{#1}}

\newcommand{\U}{\mathbf{u}}

\usepackage[pagebackref,breaklinks,colorlinks,citecolor=eccvblue]{hyperref}

\usepackage{orcidlink}

\begin{document}

\title{Event-Based Motion Magnification} 

\titlerunning{Event-Based Motion Magnification}


\author{Yutian Chen\inst{1,2^*} \and
Shi Guo\inst{1^*,\dagger} \and
Fangzheng Yu\inst{1,3} \and  Feng Zhang\inst{1} \and  \\
 Jinwei Gu\inst{2} \and Tianfan Xue\inst{2}}

\renewcommand{\thefootnote}{\fnsymbol{footnote}}
\footnotetext[0]{\hspace{-1.1em} \textsuperscript{*} indicates equal contributions. This work was done during Yutian Chen's internship at Shanghai Artificial Intelligence Laboratory.}
\footnotetext[0]{\hspace{-1.1em} \textsuperscript{\dag} indicates corresponding author.}

\authorrunning{Yutian Chen, Shi Guo, et al.}



\institute{$^1$Shanghai AI Laboratory, 
$^2$The Chinese University of Hong Kong, and \\
$^3$Zhejiang University\\
}
\maketitle

\begin{abstract}
  Detecting and magnifying imperceptible high-frequency motions in real-world scenarios has substantial implications for industrial and medical applications. These motions are characterized by small amplitudes and high frequencies. Traditional motion magnification methods rely on costly high-speed cameras or active light sources, which limit the scope of their applications. In this work, we propose a dual-camera system consisting of an event camera and a conventional RGB camera for video motion magnification, providing temporally-dense information from the event stream and spatially-dense data from the RGB images. This innovative combination enables a broad and cost-effective amplification of high-frequency motions. By revisiting the physical camera model, we observe that estimating motion direction and magnitude necessitates the integration of event streams with additional image features. On this basis, we propose a novel deep network tailored for event-based motion magnification. Our approach utilizes the Second-order Recurrent Propagation module to proficiently interpolate multiple frames while addressing artifacts and distortions induced by magnified motions. Additionally, we employ a temporal filter to distinguish between noise and useful signals, thus minimizing the impact of noise. We also introduced the first event-based motion magnification dataset, which includes a synthetic subset and a real-captured subset for training and benchmarking. Through extensive experiments in magnifying small-amplitude, high-frequency motions, we demonstrate the effectiveness and accuracy of our dual-camera system and network, offering a cost-effective and flexible solution for motion detection and magnification. Project website: \href{https://openimaginglab.github.io/emm/}{https://openimaginglab.github.io/emm/}.
\end{abstract}

\begin{figure*}[!t]
    \centering
    \begin{overpic}[width=1\textwidth]{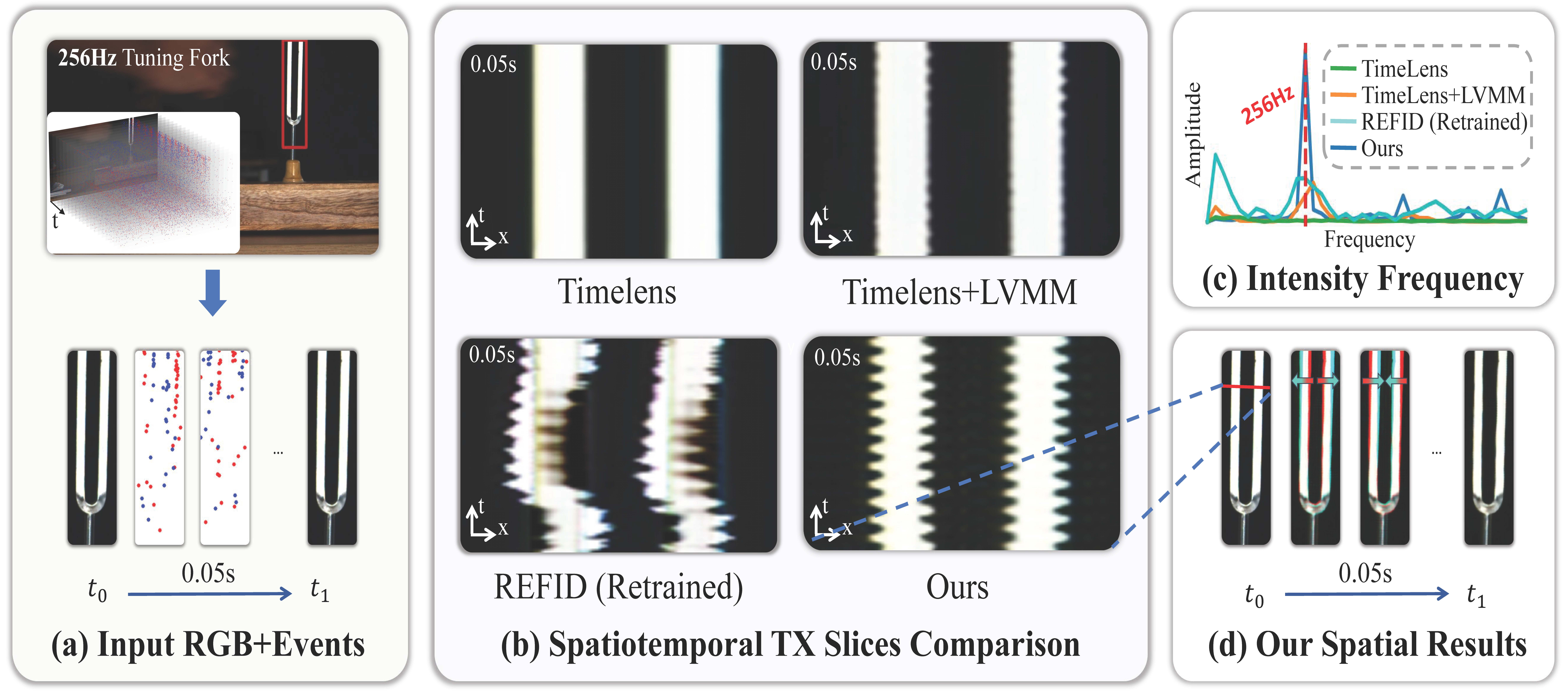}
    
    \end{overpic}
    \caption{\textbf{The magnified result of the real-captured video. Our proposed model can effectively magnify the motions at the correct frequencies compared to other solutions.} For better visualization of the tuning fork's motion, we denote different pixels between different frames using red and cyan respectively in (d). \textbf{Please refer to the supplementary videos for better visualization of results.}}
    \label{fig:teaser}
\end{figure*}

\section{Introduction}
\label{sec:intro}

In the real world, numerous small motions often escape human perception but hold significant implications for industrial and medical applications. These motions are characterized by small amplitudes, ranging from micrometers to meters, and high frequencies, spanning from a few Hz to MHz. Motion magnification technology facilitates the visualization of these imperceptible motions, thus enabling us to detect object vibrations~\cite{liu2005motion,oh2018learning}, remotely measure an individual's vital signs~\cite{wu2012eulerian}, and analyze the physical properties of materials~\cite{bouman2013estimating,davis2015visual,buyukozturk2016smaller}. Such capabilities open doors to a wide array of practical applications.

In high-frequency small motion magnification, spectral aliasing occurs if the camera system's Nyquist frequency is lower than the motion's highest inherent frequency. This limitation makes conventional camera systems impractical for such applications. Therefore, previous image-based motion magnification methods~\cite{liu2005motion,wu2012eulerian,wadhwa2013phase,wadhwa2014riesz,zhang2017video,oh2018learning,DeepMag, singh2023multi} have traditionally relied upon specialized high-speed cameras. However, these high-speed cameras are characterized by their high cost, substantial memory requirements, and limited spatial resolution. In response to the constraints posed by high-speed cameras, Sheinin \etal~\cite{sheinin2022dual} introduced a dual-shutter system. Nevertheless, this method necessitates an active light source and is primarily suited for motion detection within a limited region, thus constraining its applicability. Therefore, efficiently and conveniently detecting and magnifying low-amplitude high-frequency motion is still challenging.

To overcome temporal frequency constraints, we introduce a dual-camera system comprising an event camera and a conventional RGB camera. Event cameras~\cite{serrano2013128,brandli2014240} are neuromorphic sensors designed to only record significant pixel-wise brightness changes. They can record event streams with remarkable temporal precision, often operating at the microsecond level. Unlike high-speed cameras, these brightness-change events are spatially sparse, and thus high-speed recording does not introduce substantial data transmission overhead, making event cameras both cheap and efficient for recording high-speed motion. This is particularly advantageous for motion magnification, where very few pixel changes occur in the presence of small motion, resulting in even sparser events. Thus, in our dual-camera system, event camera captures temporally-dense information, complemented by the spatially-dense data provided by the conventional RGB camera (as shown in~\cref{fig:teaser}(a)). Such combination enhances our system's capability in detecting and amplifying non-linear high-frequency motions. 

However, employing our dual-camera system to address the motion magnification presents numerous challenges. Firstly, the task of amplifying high-frequency, sub-pixel motion results in sparse event data, which can easily be affected by noise from the event camera~\cite{hu2021v2e,rios2023within,gracca2023shining}. Second, to visualize imperceptible high-frequency motion, it is crucial to leverage the event signal to increase the frame rate of RGB images by approximately 100 times and to magnify the motion amplitude by 30 to 80 times. 
 These requirements significantly exceed the capabilities of previous event-based interpolation networks~\cite{tulyakov2021time,tulyakov2022time,he2022timereplayer,sun2023event}. As shown in \cref{fig:teaser}(b), even though the state-of-the-art event-based interpolation algorithm REFID~\cite{sun2023event} is retrained on event-based magnification data, it contains severe artifacts and distortions.
Moreover, as the first attempt to use event and RGB images for motion magnification, there are no suitable training data. Thus, designing the algorithmic solution and construct datasets for event-based motion magnification still remain challenging and open questions.

To address these challenges, we first revisit physical RGB and event camera models and derive an explicit form for the solution. 
The solution reveals a simple relationship between tiny motion to be magnified, local RGB image features, and polarities of event streams. Based on this key observation, we propose a novel deep network tailored for event-based motion magnification. Our network comprises: 1) an encoder that extracts texture, shape, and motion representations; 2) a manipulator that amplifies the motion; and 3) a decoder that reconstructs high-frame-rate, motion-magnified RGB videos. In the encoder stage, to address the challenge of interpolating a large number of frames and minimize artifacts, the features are extracted with a Second-order Recurrent Propagation (SRP) strategy. The SRP provides a better ability to model long-term information compared to previous event-based interpolation~\cite{tulyakov2021time,tulyakov2022time,he2022timereplayer,sun2023event} and motion magnification methods~\cite{oh2018learning}. In the manipulation stage, unlike conventional event-based interpolation methods~\cite{sun2023event,tulyakov2021time,tulyakov2022time}, we extract a implicit motion representation with a linear relationship to the motion's magnitude that possesses physical significance. This extraction allows for the application of the temporal filter~\cite{wadhwa2013phase,oh2018learning} to the motion representation during the testing phase, effectively filtering out noise and isolating motion information for clearer motion magnification. Experiments demonstrate that our method delivers promising performance on event-based motion magnification.

Furthermore, to provide training data for event-based motion magnification, we have developed the first dataset containing both event and RGB signals with sub-pixel motion, denoted as SM-ERGB (Sub-pixel Motion Event-RGB dataset). This dataset contain both a synthetic subset (SM-ERGB-Synth) and a real-captured subset (SM-ERGB-Real), aiming to establish a benchmark for the field. In creating SM-ERGB-Synth, we referred to the method in LVMM~\cite{oh2018learning} for blending foreground and background scenes, constructing RGB sequences that contain sub-pixel motion along with their corresponding ground-truth magnified-motion sequences. The event signals within SM-ERGB-Synth are generated using the V2E event simulator~\cite{hu2021v2e}. To validate our methods in real-world scenarios, we have also developed a dual-camera system for capturing high-quality real-world testset. The SM-ERGB-Real dataset focuses on scenes demonstrating small, high-frequency motions, \eg, a 512Hz tuning fork. Our experimental validation demonstrates that methods trained on the simulated data from SM-ERGB exhibit robust generalization to real-world scenes, effectively and accurately recovering the inherent frequencies of objects.



In summary, our contributions are as follows:
\begin{enumerate}
    \item \textbf{Novel computational imaging system for motion magnification.} To our knowledge, it is the first system that utilize temporally-dense event streams and spatially-dense RGB images to magnify imperceptible high-frequency motion. This system enables a broad and cost-effective amplification of high-frequency motions.
    \item \textbf{Novel physics-based network.} We design a novel network tailored for event-based video motion magnification, based on formation model of the imaging system. By using Second-order Porpagation and temporal filter, our model well recovers high-frequency magnified motion.  
    \item \textbf{Novel dataset.} We introduced the first event-based motion magnification dataset, Sub-pixel Motion Event and RGB (SM-ERGB) dataset, containing both synthetic (10,000 sequences) and real-captured (7 sequences) data. The dataset can promote the field of event-based motion magnification.
\end{enumerate}

\section{Related Work}
\label{sec:rela}

\subsection{Video Motion Magnification}
To detect and magnify high-frequency motion, previous video motion magnification methods~\cite{liu2005motion,wu2012eulerian,wadhwa2013phase,wadhwa2014riesz,zhang2017video,oh2018learning} have primarily relied on expensive 2D RGB high-speed cameras. These methods can be categorized into two main groups: Lagrangian-based and Eulerian-based approaches. Lagrangian methods~\cite{liu2005motion} explicitly estimate motion and warp video frames based on magnified velocity vectors. However, motion estimation in such methods is challenging and computationally demanding, often leading to visible errors in the results. In contrast, Eulerian approaches~\cite{wu2012eulerian,wadhwa2013phase,wadhwa2014riesz,oh2018learning,zhang2017video} decompose video frames into representations using Laplacian pyramids~\cite{wu2012eulerian}, complex steerable pyramid~\cite{wadhwa2013phase,wadhwa2014riesz} and deep network~\cite{oh2018learning}, and then amplify temporal variations at each pixel to magnify motion. However, the 2D RGB high-speed cameras are very expensive, bulky, require extensive data storage capacity, and have lower spatial resolution. To address the limitations associated with high-speed cameras, 
Sheinin \etal~\cite{sheinin2022dual} introduced a dual-shutter system that employs two standard cameras (60 and 134 Hz) to capture high-frequency motion, reaching frequencies of up to 63 kHz. This approach leverages a rolling shutter sensor for high-speed 1D motion capture and a global rolling shutter sensor to recover high-frequency 2D shifts within the rows of the rolling shutter sensor. However, this method relies on an active light source, which imposes constraints on its application scenarios. To overcome the limitations of previous methods, we utilize the dual camera with the temporally-dense event signals and the spatially-dense RGB images to cost-effectively magnify high-frequency motion.

\subsection{Event-based High-speed Scene Imaging}
Event cameras~\cite{serrano2013128,brandli2014240} represent a novel class of neuromorphic sensors that asynchronously obtain per-pixel intensity changes whenever the logarithmic change in latent irradiance exceeds a preset threshold. In contrast to RGB cameras, event cameras offer several advantages, including high temporal resolution, and ultra-low latency on the order of microseconds, which makes event cameras particularly well-suited for supporting high-speed imaging applications, such as frame interpolation. By leveraging event signals to interpolate high-speed scenes from low-frame-rates RGB videos, Tulyakov \etal\cite{tulyakov2021time} introduced the TimeLens framework by combining synthesis-based and flow-based methods. Building upon this, TimeLens++~\cite{tulyakov2022time} enhances the framework with multi-scale feature-level fusion and computes one-shot non-linear inter-frame motion estimation. To fully explore the potential of event cameras, He~\etal\cite{he2022timereplayer} introduced the TimeReplayer, which eliminates the need for high-speed training data by employing an unsupervised cycle-consistent strategy. Additionally, REFID~\cite{sun2023event} acknowledges that input low-frame-rate videos may suffer from blur and proposes a bidirectional recurrent network to jointly address interpolation and deblurring.

However, previous interpolation methods~\cite{tulyakov2021time,tulyakov2022time,he2022timereplayer,sun2023event} typically consider large motion and interpolate 3$\sim$15 frames. While our task requires interpolating up to 80 frames for sub-pixel motion. Thus we utilize a Second-order Recurrent Propagation strategy and temporal filter to address the challenge of long-term information modeling and noise removal.

\section{Method}


\subsection{Problem Formulation}
\label{subsec:problem_formulation}
\textbf{Event representation.} Event cameras measure brightness changes as an asynchronous stream of events. Each event is triggered when the intensity change is larger than a contrast threshold $c$. Each event $e_i$ consists of $(\mathbf{u}_i, t_i, p_i)$, where $\mathbf{u}_i = (x_i, y_i)$ are the pixel coordinates and $t_i$ is the timestamp. The event polarity $p_i$ can be expressed as:
\begin{equation}
    p_i = \begin{cases}
        1, & \text{if } \log(I_{t_i}(\U_i)) - \log(I_{t_i-\delta t}(\U_i)) \geq c, \\ 
        -1, & \text{if } \log(I_{t_i}(\U_i)) - \log(I_{t_i-\delta t}(\U_i)) \leq -c,\\        
        0, & \text{otherwise}, 
        \end{cases}       
\label{eq:define_evs}
\end{equation}
where $I_{t_i}(\mathbf{u}_i)$ is the image intensity at time $t_i$ in location $\mathbf{u}_i$, $c$ is the threshold, $\delta t$ is the time interval since the last event occurred at $\mathbf{u}_i$.

\noindent\textbf{Event-based video motion magnification.} The input of our task is two RGB frames $I_{0}$ and $I_{1}$ at timestamp $t_0$ and $t_1$, the corresponding asynchronous events $\{e_i = (\mathbf{u}_i, t_i, p_i)\}_{i=t_0}^{t_1}$ and the adjustable magnification factor $\alpha$. Our goal is to visualize and magnify the high-frequency motion between $t_0$ and $t_1$, and obtain the motion magnified frames $\{\hat{I}_i\}_{i=t_0}^{t_1}$. The number of interpolation frames $\hat{I}$ can be customized based on the motion frequency. 

To solve the above problem, we derive the formula for motion amplification using low-frame-rate images and corresponding events. Following \cite{wadhwa2013phase, oh2018learning}, the interpolated un-magnified frame $I_{\tau}$ at timestamp $\tau$ can be expressed as:
\begin{equation}
    I_{\tau}(\mathbf{u}) = I_0(\mathbf{u} - \delta_{\tau^-}(\mathbf{u})),
\label{eq:optical_flow}
\end{equation}
where $\delta_{\tau^-}(\mathbf{u})$ represents the motion field (or optical flow) from frame $I_0$ to $I_{\tau}$.

Under the assumption that the motion vector is small and brightness remains constant, we can use the first-order Taylor expansion to approximate~\cref{eq:optical_flow} as:
\begin{equation}
    I_\tau(\mathbf{u}) \approx I_0(\mathbf{u}) - I'_{0}(\mathbf{u})\delta_{\tau^-}(\mathbf{u}),
\label{eq:taylor}
\end{equation}
where $I'(\mathbf{u})$ is the image gradient. As the definition of event in~\cref{eq:define_evs}, the interpolated frame can be also modeled using event stream by using~\cite{jiang2020learning,sun2022event,lin2020learning}:
\begin{equation}
    I_\tau(\mathbf{u}) = I_0(\mathbf{u})\exp\left(c\int_{t_0}^{\tau} p(\mathbf{u},t) \, dt\right).
\label{eq:optical_event}
\end{equation}
Thus the motion field  $\delta_{\tau^-}(\mathbf{u})$ can be obtained by solving \cref{eq:taylor,eq:optical_event}. For better comprehension, we provide a 1D solution as:
\begin{equation}
    \delta_{\tau^-}(\mathbf{u}) = \frac{\sum_{\mathbf{u}\in P }c \cdot s(\mathbf{u}) \cdot (\int_{t_0}^{\tau} p(\mathbf{u},t) \, dt) }{\sum_{\mathbf{u}\in P} s^2(\mathbf{u})},
\label{eq:solve_motion}
\end{equation}
where $s(\mathbf{u}) = - I'_0(\mathbf{u}) / I_0(\mathbf{u})$ and $P$ is a small window centered at $\mathbf{u}$. Detailed derivations of 1D and 2D solutions are in the \textbf{supplementary material}.

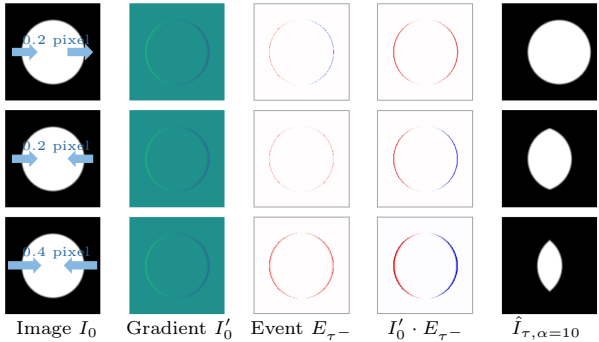
\begin{figure*}[!t]
    \centering
    \begin{overpic}[width=0.65\textwidth]{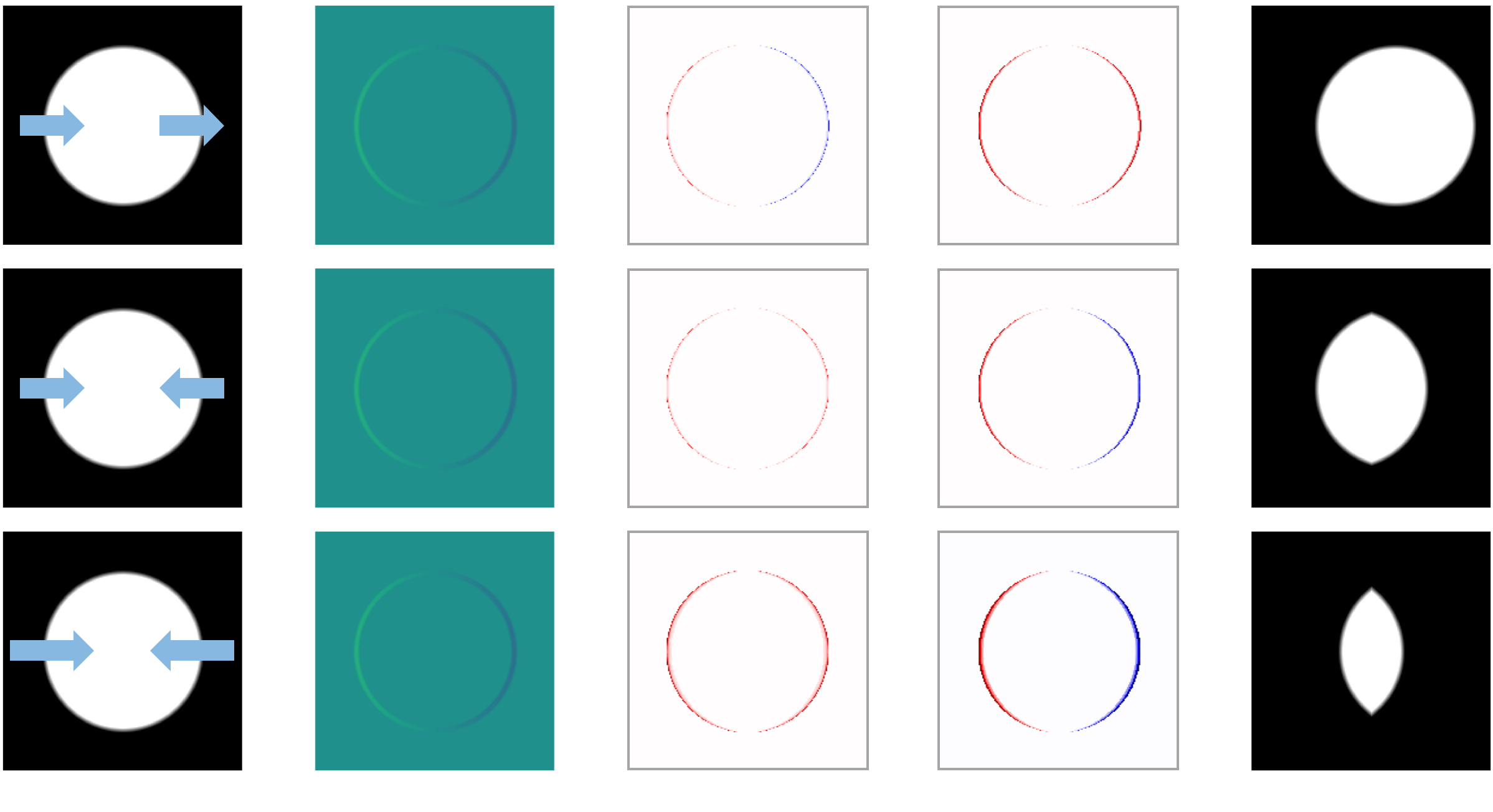}
    \put(2,-0.6){\color{black}{\scriptsize Image $I_{0}$}}
    
    \put(3,47.5){\textcolor[RGB]{38,104,155}{\tiny $0.2$ pixel}}
    \put(3,29.9){\textcolor[RGB]{38,104,155}{\tiny $0.2$ pixel}}
    \put(3,12.35){\textcolor[RGB]{38,104,155}{\tiny $0.4$ pixel}}
    
    \put(20.5,-0.6){\color{black}{\scriptsize Gradient $I'_{0}$}}
    
    \put(41.5,-0.6){\color{black}{\scriptsize Event $E_{\tau^-}$}}
    \put(64.5,-0.6){\color{black}{\scriptsize $I'_0 \cdot E_{\tau^-}$}}
    \put(85.5,-0.6){\color{black}{\scriptsize $\hat{I}_{{\tau},\alpha=10}$}}
    \end{overpic}
    \caption{\textbf{A toy example of 1-D sub-pixel motions of a circular object, demonstrating the connection between images $I_0$ (and its gradients $I'_0$), aggregated events $E_{\tau^-}$ from time stamp $t_0$ to $\tau$(blue positive, red negative), and magnified motion $\hat{I}_{{\tau},\alpha=10}$}. We can observe that event polarity itself cannot indicate the motion direction, as the left side and right side of the circular object in the first row move to the same direction, but generate events of different polarities (column 3). However, the multiplication of image gradient and event polarity (column 4, denoted as $I'_{0} \cdot E_{\tau^-}$ ) actually correlates with the motion direction. Because on both sides, the multiplication is of the same polarity, and the object is also moving in the same direction. In addition, the third row illustrates a scenario where the image gradients and motion direction are consistent with the second row, but the displacement increases from 0.2 to 0.4. This results in a larger number of generated events compared to the second row, hence amplifying the perceived magnitude of motion.}
    \label{fig:toy_example}
\end{figure*}

For an intuitive visualization of the relationship between images, events, and motions in \cref{eq:solve_motion}, we employ a 1D sub-pixel motion of a circular object as a toy example in \cref{fig:toy_example}. It underscores the necessity of using both image and event information to deduce the direction and magnitude of motion accurately, which is consistent with \cref{eq:solve_motion}.

As the motion field $\delta_{\tau^-}(\mathbf{u})$ can be solved using both RGB and Event signal as \cref{eq:solve_motion}, the magnified frame $\hat{I}_{\tau}$ can be expressed as:
\begin{align}
    \hat{I}_\tau(\mathbf{u}) &= I_0 (\mathbf{u} - (1 + \alpha) \delta_{\tau^-}(\mathbf{u})).
\label{eq:solve_of_mag_frame}
\end{align}
One can observe that, in contrast to previous methods~\cite{wu2012eulerian,wadhwa2013phase,wadhwa2014riesz,oh2018learning,zhang2017video}, instead of relying on high-frame-rate RGB frames, the proposed approach leverages the temporally-dense event stream and spatially-dense image content information for motion estimation, thus obviating the need for costly high-speed RGB cameras.

\subsection{Network Architecture}

\begin{figure*}[!t]
    \centering
    \begin{overpic}[width=1\textwidth]{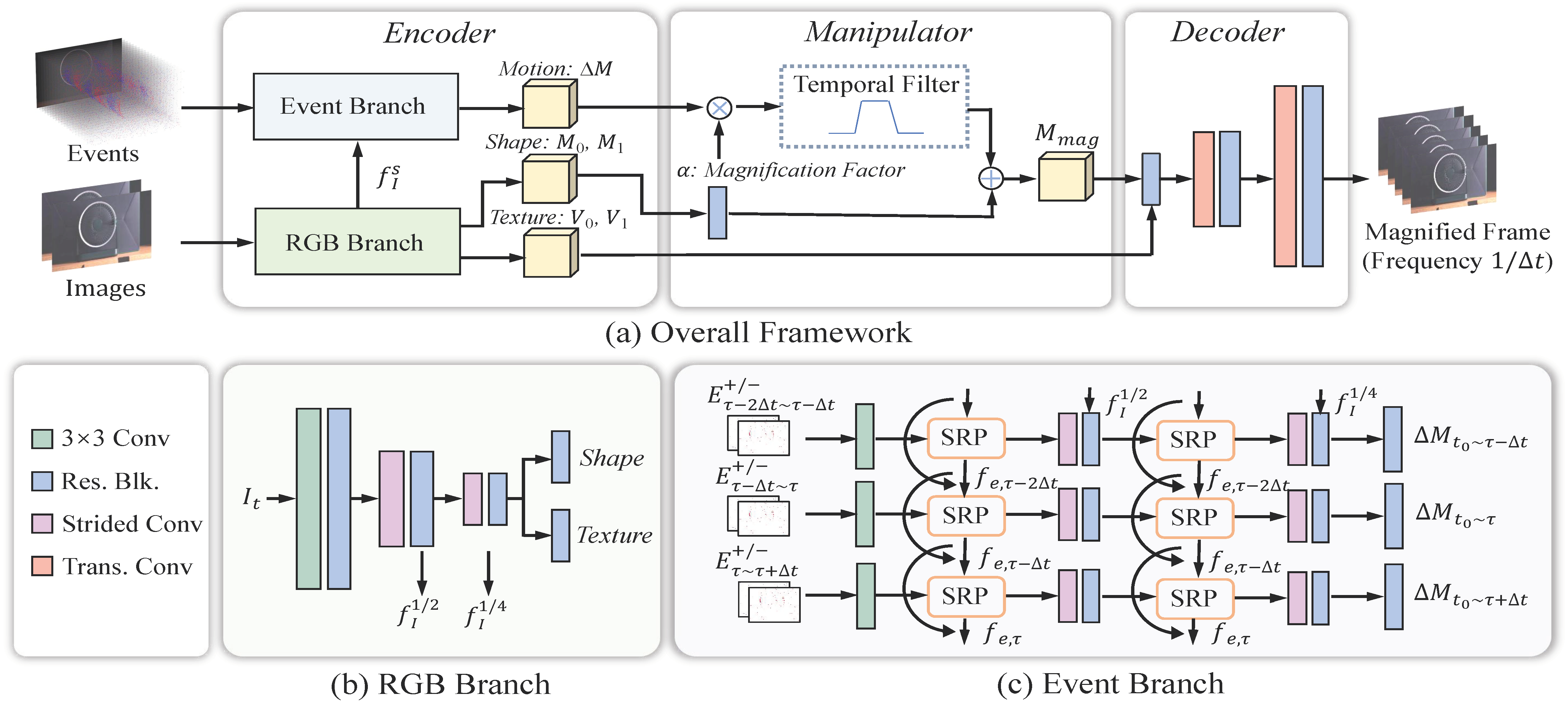}
    \end{overpic}

    \caption{\textbf{Illustration of the framework of our network.} The network is designed to utilize RGB images as well as the corresponding asynchronous events stream to generate high frame rate magnified frames. (a) is the overall framework, which consists of three main components: Encoder, Manipulator, and Decoder. In the Encoder, the RGB Branch (b) extracts temporal-invariant texture representations $V_{0}$ and $V_{1}$, as well as shape representations $M$, while the Event Branch (c) extracts temporal-variant motion representations $\Delta M$. The Manipulator linearly magnifies the motion using factor $\alpha$, producing the magnified motion $M_{mag}$. To amplify motion at specific frequencies and reduce noise interference, a temporal filter is employed during inference. The Decoder then reconstructs motion-magnified frames leveraging $M_{mag}$ and texture data.}
    \label{fig:main-network}
\end{figure*}

Although the derivation in \cref{subsec:problem_formulation} mathematically formulates the relation among magnified motion, image, and event streams, directly utilizing this results in dual-stream motion magnification may result poor performance, as it based on some strong assumptions, \eg, brightness constancy.

Therefore, we abstract the physical model in \cref{eq:solve_of_mag_frame} into an end-to-end neural network to learn abstractions and priors from data, enhancing its adaptability across diverse and complex scenarios. The overview of our network is illustrated in Fig.~\ref{fig:main-network}. Inspired by~\cite{wadhwa2013phase,oh2018learning}, Our network consists of three parts: the encoder for representation extraction, the manipulator for magnifying motion, and the decoder for frame reconstruction. 

\textbf{Encoder.} The encoder is comprised of two branches: the image branch and the event branch. The image branch is dedicated to extract texture representation ($V_{0}$ and $V_1$) and shape representation ($M_{0}$ and $M_{1}$) from the RGB images ($I_{0}$ and $I_{1}$). The event branch is used to extract motion representation $\Delta M_{\tau}$ between $t_0$ and any interval time $\tau \in (t_0, t_1)$, which is an implicit representation of $\delta_{\tau^-}(\mathbf{u})$ in ~\cref{eq:solve_of_mag_frame}. 

For the RGB branch, the RGB images are firstly modeled into deep features. To reduce the effect of noise and save computational cost, feature map $f_{t}$ uses $1/8$ resolution compared with the input. Then the texture representations $V_t$ and shape  representations $M_t$ are extracted by the corresponding network. Note that $(M_{0} - M_{1})$ represents the shape change between two RGB frames.


For the event branch, the events are first converted into a voxel grid to enable neural networks to process~\cite{tulyakov2021time,tulyakov2022time,sun2023event}, denoted as $E\in \mathbb{R}^{2n\times H\times W}$, where $n$ is the interpolated frame number. Following \cite{gehrig2019end}, we accumulate the events in a small time interval $\Delta t$  into two distinct voxel grids $E^{+}$ and $E^{-}$ for positive and negative events, respectively. Such a method can prevent filtering out small motion caused by directly accumulating positive and negative events together and retain a more comprehensive set of information.

To capture high-frequency shape changes between $t_0$ and $t_1$, both the event stream and RGB features are employed, as stated in~\cref{eq:solve_motion}. Similarly, directly extracting the motion field $\delta_{\tau^-}(\mathbf{u})$ using~\cref{eq:solve_motion} may face some robustness issues. Instead, we follow the Eulerian approaches~\cite{wu2012eulerian,wadhwa2013phase,wadhwa2014riesz,oh2018learning,zhang2017video} and extract the features' implicit motion $\Delta M_{\tau}$ from deep images features $f_{I}$ and the events between $t_0$ and $\tau$ using neural network $F_{\Delta M}$.
%
$F_{\Delta M}$ consists of strided convolutions, residual blocks, channel-attention fusion (CAF) module, and our proposed second-order recurrent propagation (SRP) module. The CAF module is used to adaptively fuse information from RGB images and event stream, inspired by \cite{sun2023event}. The proposed SRP module can aggregate information over long-term temporal events, and this enhancement is particularly beneficial for high-frequency motion magnification. \cref{subsec:second-order} will further discuss its detailed implementation.

\textbf{Manipulator.} The event branch of encoder stated above can extract the motion information, represented the change of shape representation $\Delta M$, from both deep RGB features and event streams. The objective of manipulator is then to generate the magnified motion representation.

One key observation of the phase-base motion magnification~\cite{wadhwa2013phase} is that when motion is small, the filter response, which is the motion representation in our setup, is linear to the actual underlying motion $\delta_\tau^-(\mathbf{u})$. 
Therefore, we directly apply the magnification factor $\alpha$ to initial motion presentation $\Delta M$, which should correspond to the motion representation for the magnified motion $\alpha \cdot \delta_\tau^-(\mathbf{u})$.
Formally, at timestamp $\tau$, the  shape feature for the magnified motion $M_{mag,\tau}$ is:
\begin{equation}
  M_{mag,\tau} = M_0 + \alpha \cdot \Delta M_{\tau}.
\label{eq:manipulator}
\end{equation}
where $M_{0}$ is the shape feature extracted from the first RGB image.

In practice, when the motion is tiny, the extract motion representation from input is noisy, and the magnification further boosts the noise. Therefore, it is often desirable to amplify motion at specific frequencies to reduce the noise. Therefore, we utilize temporal filters~\cite{wadhwa2013phase,oh2018learning} into the manipulator in the inference phase, as $\hat{M}_{mag,\tau} = M_0 + \alpha \cdot \Gamma (\Delta M_{\tau})$, where $\Gamma(\cdot)$ is a 1D temporal band-pass filter. To apply this temporal filter, we apply pixel-wise Fourier transformation on per-frame motion presentations $\Delta M_{\tau}$ along the temporal dimension, extract the desired frequency of motion using a band-pass filter, and apply an inverse Fourier transformation to return to the time domain.

\textbf{Decoder.} The decoder reconstructs magnified frames $\{\hat{I}_t\}_{i=t_0}^{t_1}$ using magnified shape representation $M_{mag}$ and the texture representation $V$ of two frames. Ideally, $V_{0}$ and $V_{1}$ represent the same information, therefore, for globally invariant texture information, we simply use the average $(V_{0}+V_{1})/2$ as input.

\subsection{Second-order Recurrent Propagation Module}
\label{subsec:second-order}
\begin{figure*}[!t]
    \centering
    \begin{overpic}[width=1\textwidth]{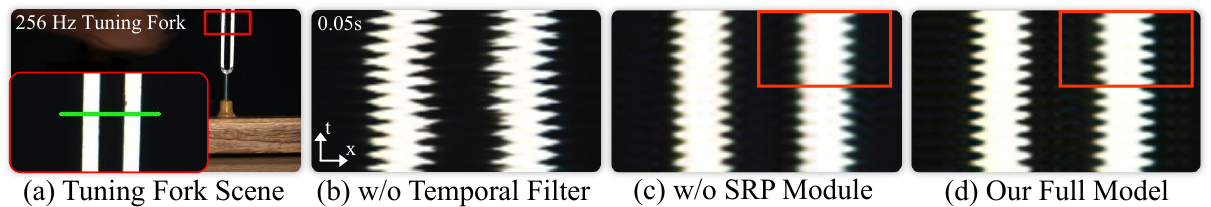}
    \end{overpic}

    \caption{Comparison of the variant of our model.}
    \label{fig:ablation_visual}
\end{figure*}

To aggregate temporal event information, there are two typical solutions, \ie, unidirectional recurrent propagation~\cite{peng2023get} and bidirectional recurrent propagation~\cite{sun2023event}. Given that the motion we need to amplify $\Delta M_{\tau}$ is computed from $t_0$ to the current timestamp $\tau$, a bidirectional approach is not well-suited to our specific problem. While  Recurrent Neural Network (RNN) structures are known for their efficiency, it might not be ideal for our scenario, since our task demands the modeling of distant temporal dependencies due to the insertion of a large number of frames, up to 100 frames. Therefore, we propose a Second-order Recurrent Propagation (SRP) Module, the effect of which is shown in~\cref{fig:ablation_visual}.

At time $\tau$, the event feature, represented as $f_{e, \tau}$, is fused with forward information $h_t$ from the two preceding event frames as:
\begin{equation}
    f{'}_{e, \tau}, h_{\tau} = F_{SRP}(f_{e, \tau}, h_{\tau-\Delta t}, h_{\tau-2\Delta t}),
\end{equation}
where $F_{SRP}$ is the SRP unit. In $F_{SRP}$, the channel attention~\cite{hu2018squeeze,zamir2020learning} is used to adaptive merge event temporal information. 


\subsection{Training Loss}
\label{subsec:loss}
At last, we design the training loss below. First, we define the reconstruction loss between the network prediction and ground-truth as $\mathcal{L}_{r} = \frac{1}{n}\sum_{t} \sqrt{\Vert \hat{I}_t - \tilde{I}_t \Vert^2 + \epsilon^2}$,
where $\tilde{I}_{t}$ is the ground-truth magnified frame, $\hat{I}_t$ is the network prediction, $n$ is the number of the interpolated frames, $\sqrt{\Vert \hat{x} - x \Vert^2 + \epsilon^2}$ is the Charbonnier penalty function, and $\epsilon$ is set to 0.001. In the \cref{sec:dataset} below describes the details of obtaining ground-truth magnification results. 

To further improvement the robustness, we use additional texture and shape losses. First, since the texture information extracted from input images $I_{0}$ and $I_{1}$ should be same, we define the texture loss as~\cite{oh2018learning} $\mathcal{L}_{t} = \sqrt{\Vert V_{0} - V_{1} \Vert^2 + \epsilon^2}$.
We also introduce a shape loss to ensure the consistency between the motions extracted from images and the event, denoted as $\mathcal{L}_{s} = \sqrt{\Vert M_{1} - M_{0} - \Delta M_{1} \Vert^2 + \epsilon^2}$.
To sum up, the overall loss to optimize our model is $\mathcal{L} = \mathcal{L}_{r} + \lambda_1 \mathcal{L}_{t} + \lambda_2 \mathcal{L}_{s}$, where $\lambda_1$ and $\lambda_2$ are trade-off parameters and we set to 0.01 in our experiment.

\subsection{SM-ERGB Dataset}
\label{sec:dataset}
Since this is the first event-RGB-based motion magnification, there is no existing dataset designed for this task. Therefore, we construct a dataset containing both event and RGB signals with sub-pixel motion, denoted as SM-ERGB, to train and evaluate our network. This dataset also aims to establish a benchmark for this field. SM-ERGB consists of two subsets. \emph{SM-ERGB-Synth} is a synthetic dataset with event and RGB signals, as well as the ground truth magnification output, which is used for training and evaluation. \emph{SM-ERGB-Real} is a real-world dataset captured by RGB-Event dual camera system. This dataset only contains input and is used for evaluation. Details of each dataset are described below.

\textbf{SM-ERGB-Synth:} 
Our simulation pipeline begins with the initial generation of videos containing small-motion and magnified-motion sequences respectively. This is achieved by compositing foreground elements onto a background, where both trajectories and magnification factors are randomly generated. Subsequently, we generate the event data using the event simulator V2E~\cite{hu2021v2e}. Following Oh \etal~\cite{oh2018learning}, we use DIV2K dataset~\cite{agustsson2017ntire, Agustsson_2017_CVPR_Workshops} for background and PASCAL VOC dataset~\cite{pascal-voc-2010,pascal-voc-2011} for foreground. To handle the magnification of motion with small amplitudes, it is important to simulate sub-pixel motions into our dataset. Therefore, we first generate high-resolution video streams, where the motion is more pronounced, and subsequently downsample each frame to the desired resolution. Specifically, we downsample the original $4096\times4096$ images to $256\times256$ images, achieving a motion as small as 0.0625 pixels. After using V2E~\cite{hu2021v2e} for simulating event data from downsampled videos, we acquire the training dataset.

In the training phase, we utilize the initial and terminal frames of the video in conjunction with the simulated event data as inputs. We employ the magnified video as the ground-truth data to offer supervision. The SM-ERGB-Synth dataset is separated into training and testing parts, with 10,000 and 50 scenes respectively.The dataset features magnification factors ranging from 30 to 80 and comprises 30 frames. The training and testing datasets use unique foregrounds and backgrounds.



\textbf{SM-ERGB-Real:} To validate the effectiveness of our approach in magnifying high-frequency motion in real-world scenarios, we build a dual camera system and collect a real-world dataset of 7 sequences, denoted as SM-ERGB-Real. This system consists of a event camera ($1080\times720$) and a RGB camera ($2048\times1536$). To ensure alignment between the fields of view from both sensors~\cite{han2020neuromorphic,wang2020joint,zhou2023deblurring}, we implemented a beam splitter positioned in front of the cameras. Both cameras have undergone geometric calibration and temporal synchronization processes. The real-world test set all contains motion occurring at sub-pixel levels. Given that our method utilizes a fully convolutional layer, it can be directly applied to images of various resolutions, even when trained on $256\times 256$ patches.


\section{Experiments}
\label{Sec:Experiments}

\subsection{Training Details}
In our experiment, we train both our network and baselines on the training set of SM-ERGB-Synth. Each scene in SM-ERGB-Synth comprises 30 consecutive motion steps.
%
The primary objective of the model is to generate 30 magnified frames from the first and last frames of each scene and the events between them. 
%
%
It is worth noting that due to the utilization of a recurrent framework, our method has the capability to generate randomly interpolated frames, such as 80 frames during our inference phase. 
Our model is implemented in Pytorch. During training, we use Adam~\cite{Kingma2014AdamAM} optimizer with an initial learning rate of $1\times10^{-4}$ and weight decay coefficient of $1\times10^{-4}$ for 200k iterations.

\subsection{Baseline Methods}
Since there is no magnification algorithms designed for dual-stream RGB-event input, we try our best to adapt the existing methods to this tasks. First, we create baseline solutions by combining the state-of-the-art RGB-event interpolation algorithm with the RGB motion magnification algorithms. Specifically, we start by generating high-frame-rate RGB video from the low-frame-rate one using the interpolation algorithm, and then magnify its motion using RGB motion magnification algorithms, \ie, LVMM~\cite{oh2018learning}.
To ensure the quality of interpolated high-frame-rate video, we also tried different interpolation algorithms, including SuperSloMo~\cite{Jiang2018super} and RIFE~\cite{huang2022real} for RGB-based interpolation and  TimeLens~\cite{tulyakov2021time} for event-based interpolation. Second, to further demonstrate that our network is better suited for motion magnification tasks compared to conventional frame interpolation networks, we retrained a state-of-the-art event-based frame interpolation network REFID~\cite{sun2023event} using our dataset to adapt it to the motion magnification task.

\subsection{Results on Synthetic Videos}

\begin{table*}[t] 
    \caption{\textbf{Quantitative comparison on SM-ERGB-Synth test set.} Our system requires only a small data stream size yet achieves performance comparable to that of LVMM on high-speed video, which requires a data stream 5 times larger. }
    \scriptsize
    \centering
    {
    \begin{tabular}{cccccc}
    
    \toprule
    Method  & Data Size (RGB+Event) &RGB FPS & Events &PSNR  & SSIM  \\ \midrule
     LVMM~\cite{oh2018learning} on highspeed video & 3.60 M + 0M& 900Hz& \XSolidBrush           & 22.30 & 0.7309   \\
     SuperSloMo~\cite{Jiang2018super} + LVMM~\cite{oh2018learning}& 0.23 M + 0 M& 30Hz& \XSolidBrush           & 14.32 & 0.3871   \\
     RIFE~\cite{huang2022real} + LVMM~\cite{oh2018learning}& 0.23 M + 0 M& 30Hz& \XSolidBrush           & 17.09          & 0.5908 \\
    TimeLens~\cite{tulyakov2021time} + LVMM~\cite{oh2018learning} &0.23 M + 0.49 M&30Hz &\Checkmark&  17.60 & 0.5582   \\
    REFID (retrained)~\cite{sun2023event}  &0.23 M + 0.49 M&30Hz &\Checkmark&  19.64 &  0.8029   \\
    Ours                       &0.23 M + 0.49 M&30Hz&  \Checkmark & \textbf{ 22.32 }& \textbf{0.8753}   \\ \bottomrule
    \end{tabular}}
    
     \label{tab:result}
\end{table*}

\cref{tab:result} shows our quantitative results on SM-ERGB-Synth dataset. Compared to directly magnify motion in high-speed RGB videos using LVMM, our event-based method presents a notable reduction in data transferring and storage demands (merely $\sim$20\% (0.72M/3.60M) of the data volume) and also improves the visual quality. This suggests that for subtle motions, high-speed cameras may produce excessive data, while utilizing event data and low-frame-rate RGB video can achieve similarly effective results at a lower cost. The results from SuperSloMo+LVMM are suboptimal due to the linear motion assumptions these algorithms imposed and low sampling rates in the input RGB stream. Furthermore, compared with TimeLens and SuperFast for event-based frame interpolation followed by LVMM, our method demonstrates an enhancement of approximately 4.7dB. This gain can be attributed to our end-to-end network, which prevents the error accumulation in the two-stage process of interpolation followed by magnification. Compared with retrained REFID, our method utilize the well-designed manipulator to integrate magnification factor and SRP model to better model long-term information, resulting in about 2.5dB improvement. 

\begin{figure*}[!t]
    \centering
    \begin{overpic}[width=1\textwidth]{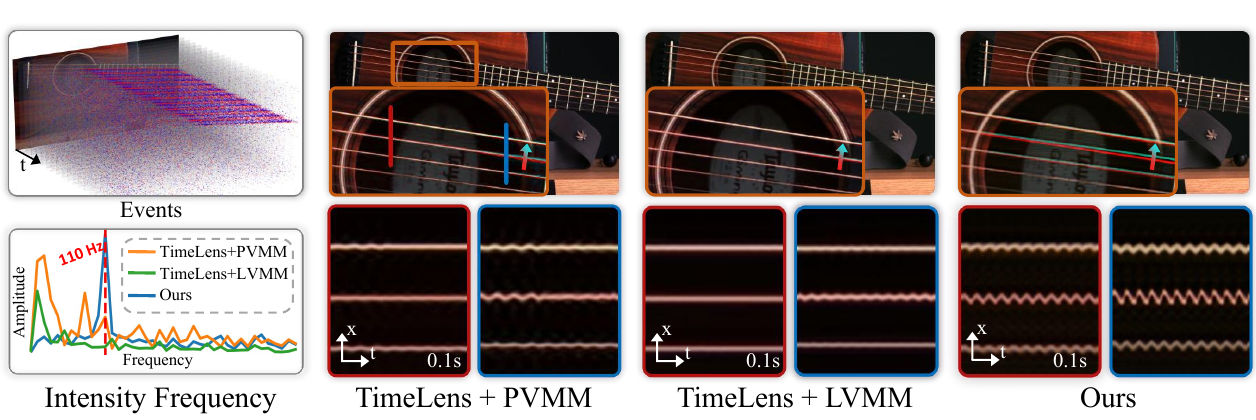}
    \end{overpic} 
    \caption{Magnification results of different methods on the real-world video of a 110Hz guitar string in SM-ERGB-Real dataset.}
    \label{fig:result_with_gt}
\end{figure*}

\begin{figure*}[!t]
    \centering

    \begin{overpic}[width=1\textwidth]{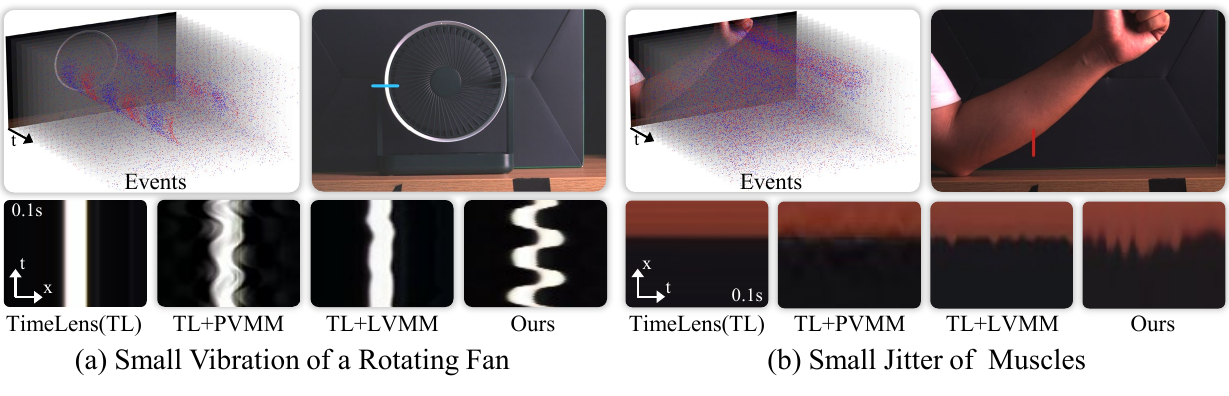}
    \end{overpic}
    \caption{Magnification results of different methods on the real-world video without ground-truth frequency in SM-ERGB-Real dataset. }

    \label{fig:result_without_gt}
\end{figure*}

\subsection{Results on Real-world Video}

We present visual comparisons on SM-ERGB-Real dataset with ground-truth frequency information, as illustrated in \cref{fig:teaser,fig:result_with_gt}. The figures depict magnified results 
from recordings of a 256Hz tuning fork and a 110Hz guitar string, respectively. It is observed that the magnification outputs derived from TimeLens+PVMM and TimeLens+LVMM fall short of representing the inherent motion frequency accurately. Following the application of the temporal filter, only a limited amount of motion can be recovered. As shown in \cref{fig:teaser}, even though the retrained REFID can magnify sub-pixel motion, it suffer from several artifact and distoration. In contrast, our proposed model effectively captures and magnifies the motions at the correct frequencies with precision. We further validate the applicability of our model to various types of motions, including scenarios where ground-truth frequency information is not available. This includes scenarios such as a small vibration
of a rotating fan and a small jitter of muscles, which is shown in \cref{fig:result_without_gt}. Compared to other methods, our approach can magnify motion in a more pronounced and rational manner. 

\subsection{Ablation Study}
\myparagraph{Temporal filter.} To evaluate the effect of the temporal filter in our model, we make a comparison on the tuning fork sequence in \cref{fig:ablation_visual} (b)(d). It is observable that even in the absence of a temporal filter, our model manages to capture motions at the correct frequencies. The temporal filter further enhances regularity in the motion by effectively decreasing the temporal noise of the event camera.

\begin{figure*}[!t]
    \centering
    \begin{overpic}[width=0.50\textwidth]{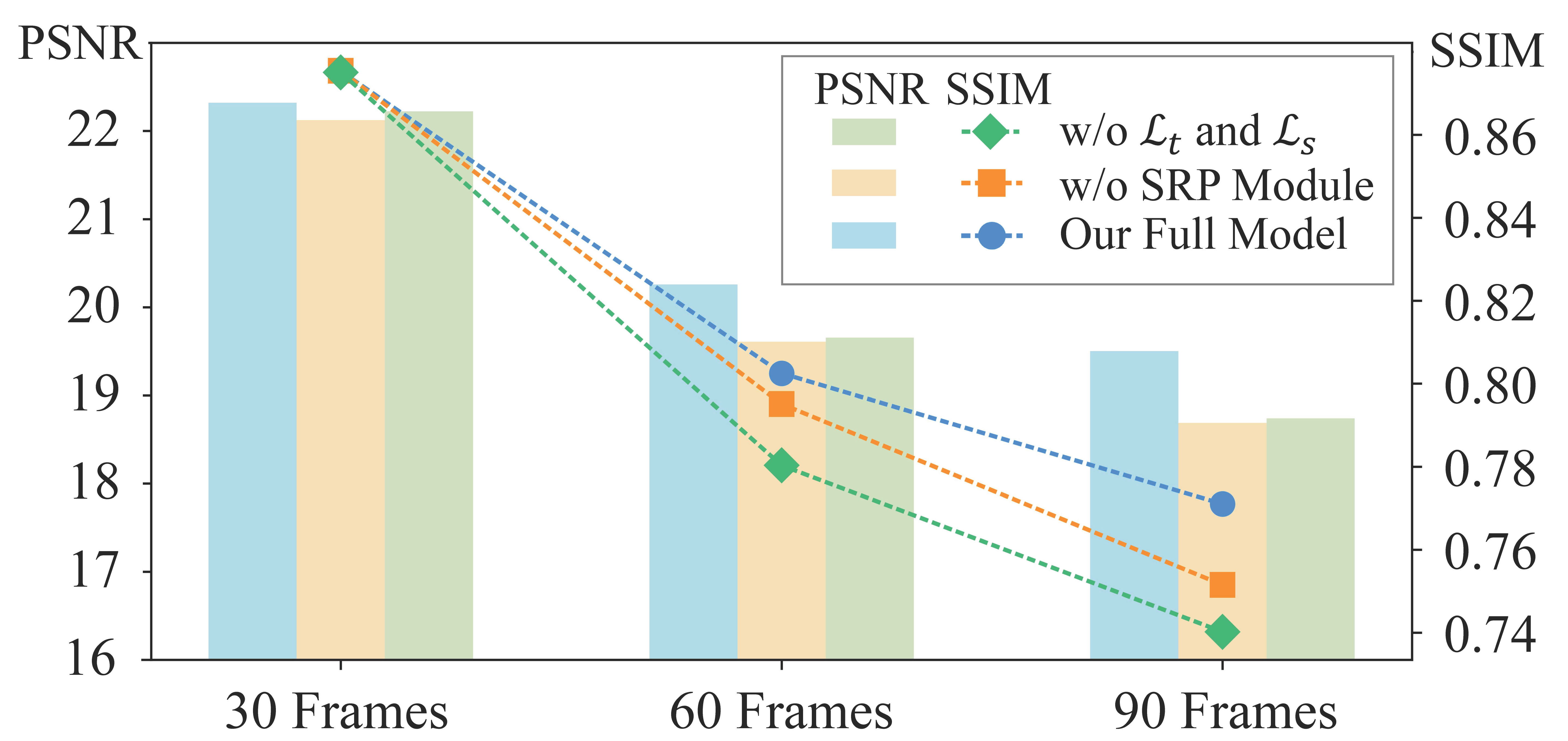}

    \end{overpic} 
        \caption{Ablation study of different architectural components.}
    \label{fig:ablation}
    
\end{figure*}

\myparagraph{SRP module.} To demonstrate the effectiveness of the SRP module, we compare it with the first-order propagation. The results are shown in \cref{fig:ablation_visual} (c)(d) and \cref{fig:ablation}. The utilization of the SRP module reduces the blurring artifacts of first-order propagation, particularly in scenarios involving a high number of frames for interpolation. \cref{fig:ablation} clearly illustrates that the improvement of the model with SRP over the one with normal RNN becomes progressively pronounced as the interpolation sequence lengthens. Both results confirm the SRP's superior performance in the preservation of long-term temporal information.

\myparagraph{Loss function.} To evaluate the performance of the loss function, we train a variant without using $\mathcal{L}_{t}$ and $\mathcal{L}_{s}$. The comparison is illustrated in \cref{fig:ablation}. The inclusion of $\mathcal{L}_{s}$ and $\mathcal{L}_{t}$ enhances the network's performance by improving its ability to utilize long-term event information and discriminate between texture and shape representations. This improvement is particularly evident when interpolating 90 magnified frames.
\section{Conclusion}
    In this paper, we addressed the challenges of magnifying high-frequency, small-amplitude motions by introducing a novel dual-camera system that consists of an event camera and a conventional RGB camera. This dual-camera system provides spatially-dense RGB information and temporally-dense event signals, offering advantages in magnifying non-linear high-frequency motions. By using the physic model of events, we provide the analytical solution for event-based motion magnification. From the insight of the solution, we proposed a novel network with the Second-order Recurrent Propagation strategy and the temporal filter. Such design effectively manages the inherent challenges in high-frequency motion magnification, \ie, interpolation of a large number of frames, the artifact and distortion caused by magnifying motions, and the blend of noise with subtle motions. Furthermore, a sub-pixel motion RGB and event dataset, SM-ERGB, is collected to establish the benchmark for event-based motion magnification. The SM-ERGB dataset contains both synthetic and real-captured part with sub-pixel motion. Through experiments, our dual-camera approach, coupled with our proposed network, has proved to be efficient and accuracy in amplifying real-world small-amplitude, high-frequency motions.

    Our method has limitations in detecting subtle motions with low contrast to the background and in flickering scenarios. These are unsolved challenges in original motion magnification as well. We plan to address this in our future work.
  
    \section*{Acknowledgements} 
    This work is partially supported by the National Key R\&D Program of China (NO.2022ZD0160201) and CUHK Direct Grants (RCFUS) No. 4055189.


%
%
\bibliographystyle{splncs04}
\bibliography{main}

\clearpage
\title{ Event-Based Motion Magnification: \\Supplementary Materials}
\authorrunning{Yutian Chen, Shi Guo, et al.}
\titlerunning{Event-Based Motion Magnification}
\author{}
\institute{}
\maketitle
In this supplementary file, we provide the following materials:

\begin{itemize}
\setlength{\itemsep}{0pt}
\setlength{\parsep}{0pt}
\setlength{\parskip}{0pt}
    \item Video comparisons of different methods;
    \item Comparisons of different interpolation method;
    \item Quantitative results on real-world data;
    \item Representation visualization (referring to Section 3.2 in the main paper);
    \item Sub-pixel motion visualization;
    \item Explicit solution of event-based motion magnification (referring to Section 3.1 in the main paper).
\end{itemize}
\setcounter{section}{0}
 \section{Video Comparisons}
We present the visual result of Figs. 1, 5 and 6 in the main paper. 
Due to the characteristic of the problem, we provide supplementary video for better visualization. The video contains video comparisions with retrained REFID~\cite{sun2023event},  Timelens~\cite{tulyakov2021time}+PVMM~\cite{wadhwa2013phase}, and Timelens~\cite{tulyakov2021time}+LVMM~\cite{oh2018learning}. One can see that our method can well reconstruct rigid body motion with accurate frequency. Additionally, the video shows results at varying magnification factors.

\section{Interpolation method.} We follow LVMM~\cite{oh2018learning} to use  bicubic downsampler, and the model trained that generalizes well to real-world scenarios.
 To provide a quantitative comparison, we evaluate the model trained on bicubic downsampling and test it on the following two different downsampling methods. \cref{table:new_comparison} shows that the interpolation methods have minimum impact to the result.

\begin{table}[h]
\setlength{\tabcolsep}{12pt}
\caption{Comparisons of different interpolation method.}
\centering
\label{table:new_comparison}
\begin{tabular}{c|c|c}
\hline
Test dataset & Lanczos downsampling          & bicubic downsampling   \\ \hline
 PSNR / SSIM&22.2402 / 0.8686 &22.2444 / 0.8669 \\\hline
 
\end{tabular}

\end{table}

\section{Quantitative results on real-world data.}
\begin{figure}[!h]
  \centering
  \includegraphics[width=0.7\linewidth]{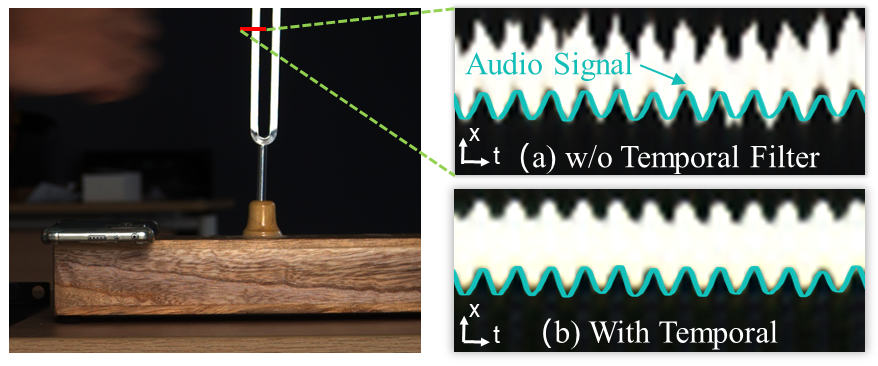}
   \caption{ \textbf{Accuracy of our method. }Comparison between our magnified output and the audio signal (blue line) over a duration of 0.05 seconds. Our result and the audio signal match closely.}
   \label{fig:sound}
\end{figure}

We used a microphone to synchronously record the audio emitted by the vibrating tuning fork and visualized the results in \cref{fig:sound}. It shows that the audio signal (blue line) aligns well with our magnified result. 

Also, we use a Relative Mean Squared Error (R-MSE) for the motion frequencies evaluation to provide quantitative comparison on real-world scenes. The observation frequency is determined by performing a Fourier transform on the time-varying intensity of the magnified moving object's edges and selecting the frequency with the highest amplitude. The results are shown in \cref{table:real_compare}, along with additional baseline. One can see that our method can well recover the natural high frequency of the objects. 



\begin{table}[h]
\centering

\centering
\caption{Quantitative comparison on real-world scenes.}
\label{table:real_compare}
\begin{tabular}{c|c|c|c}
\hline
 &SuperSloMo + LVMM& TimeLens + LVMM & Ours          \\ \hline
R-MSE &0.590 & 0.411        &\textbf{0.002} 
 \\ \hline

\end{tabular}
\end{table}

\section{Representation Visualization}

\begin{figure}[!h]
    \centering

    \begin{overpic}[width=0.7\textwidth]{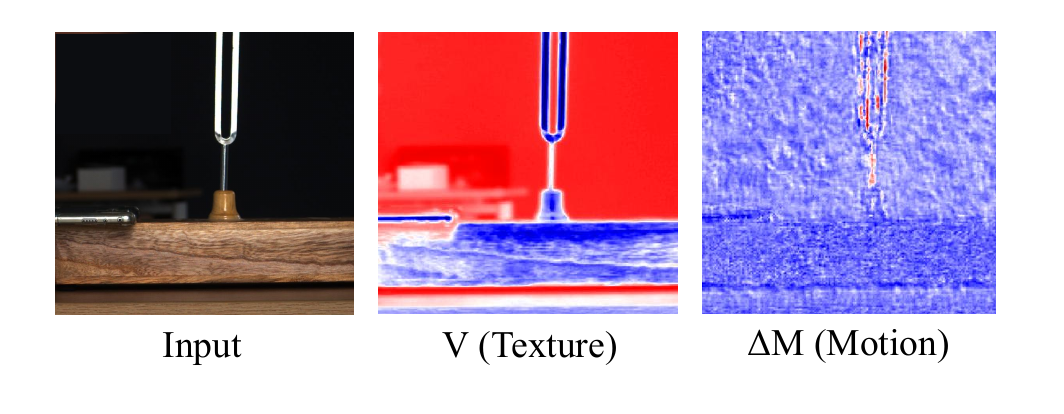}
    \end{overpic}
    \caption{A visualization of texture representation (donated as $V$) and motion representation (donated as $\Delta M$) in the network.}
    \label{fig:repre}
\end{figure}

To better interpret our network, we visualize one channel of texture representation and motion representation in our network. As we can see in \cref{fig:repre}, the texture representation captures the wood's grain and the background stationary objects while the motion representation mainly captures the moving edges of the tuning fork and the movements of the block supporting the tuning fork. This further illustrates our encoder's capability to distinguish and extract the texture information and motion information separately.

\section{Sub-pixel Motion Visualization}
\begin{figure}[!h]
    \centering

    \begin{overpic}[width=1\textwidth]{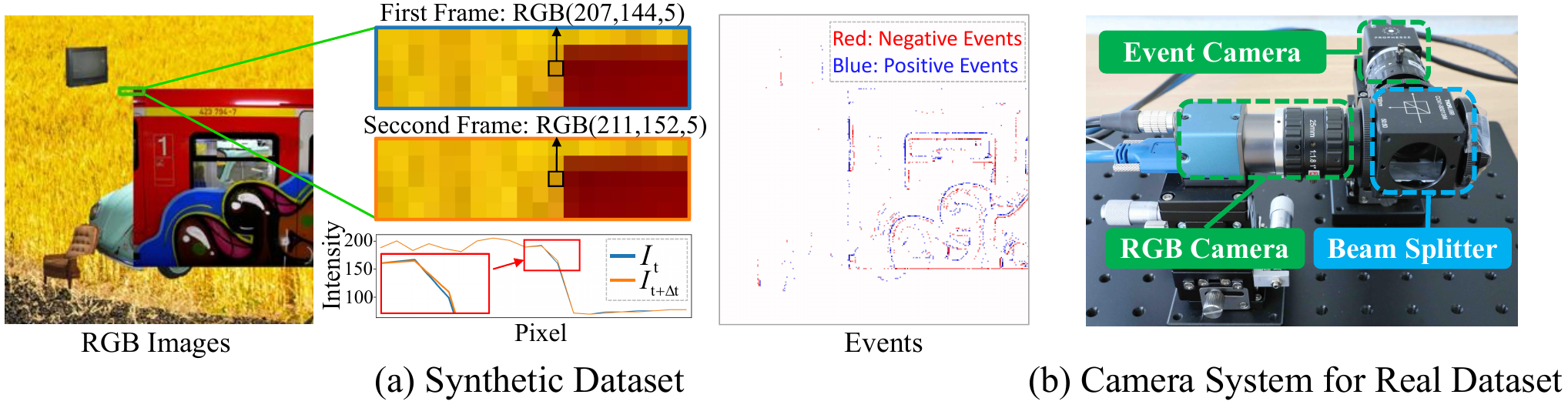}
    \end{overpic}
    \caption{(a) An example of our synthetic dataset, (b) RGB-Event dual camera system.}
    \label{fig:dataset}
\end{figure}

In \cref{fig:dataset}(a), we visualize sub-pixel motions in our synthetic dataset. It can be observed that the primary manifestation of these sub-pixel motions is in the intensity variations of RGB images. Moreover, the corresponding event signals effectively record these sub-pixel movements.

\section{Derivations of Explicit Motion Solution based on Event and RGB}

    Based on Eqs. (3) and (4)  in the main paper, we derive:
    \begin{equation}
    I(\mathbf{u}) - I'(\mathbf{u}) \delta_{\tau^-}(\mathbf{u}) = I(\mathbf{u}) \cdot \exp(c\int_{t_0}^{\tau} p(\mathbf{u},t) \, dt ),
    \end{equation}
Let us define $s(\mathbf{u})$ as $- I'(\mathbf{u}) / I(\mathbf{u})$, and divide $I(\mathbf{u})$ on both sides of the equation above, we get:
    \begin{equation}
        s(\mathbf{u})\delta_{\tau^-}(\mathbf{u}) =  \exp(c  \int_{t_0}^{\tau} p(\mathbf{u},t) \, dt) - 1 \approx c\int_{t_0}^{\tau} p(\mathbf{u},t) \, dt.
    \label{eq:supp_motion_model}
    \end{equation}
    
    \textbf{1D Solution:} Following the Lucas-Kanade method \cite{lucas1981iterative}, we assume that  the motion field (or optical flow) $\delta_{\tau^-}(\mathbf{u})$ approximately constant within a window $P$ centered at the pixel coordinate $\mathbf{u}$ and $\mathbf{u_1},\mathbf{u_2},\dots,\mathbf{u_n}$ are the pixels inside the window. The objective function of \cref{eq:supp_motion_model} is:
    \begin{equation}
        \min_\delta \| \mathbf{A}\delta_{\tau^-}(\mathbf{u}) -  \mathbf{b}\|^2,
    \end{equation}
    where $\mathbf{A} = [s(\mathbf{u_1}) \cdots s(\mathbf{u_n})]^{T}$ and $\mathbf{b} = [c   \int_{t_0}^{\tau} p(\mathbf{u},t) \cdots  c\int_{t_0}^{\tau} p(\mathbf{u},t)]^{T}$. The minimum value is:
    \begin{equation}
    \delta_{\tau^-}(\mathbf{u}) = \frac{\sum_{\mathbf{u}\in P} s(\mathbf{u})  c  \int_{t_0}^{\tau} p(\mathbf{u},t) \, dt  }{\sum_{\mathbf{u}\in P} s^2(\mathbf{u})}.
    \end{equation}
    And recall again, $s(\mathbf{u}) = -I'(\mathbf{u}) / I(\mathbf{u})$.

    \textbf{2D Solution: } Similarly, we define a window $P$ centered at the pixel coordinate $\mathbf{u}$. Compare with 1D motion, the image gradient $I'(\mathbf{u})$ and motion $\delta_{\tau^-}(\mathbf{u})$ becomes 2D vector. Thus, we define $s(\mathbf{u}) = \begin{bmatrix} -I_x(\mathbf{u})/I(\mathbf{u}) & -I_y(\mathbf{u})/I(\mathbf{u})  \end{bmatrix} =\begin{bmatrix} s_x(\mathbf{u}) &  s_y(\mathbf{u})  \end{bmatrix} $, and the objective function is:
    \renewcommand{\arraystretch}{1.3}
    
    \begin{align}
    &\qquad \min_{\delta} \|\mathbf{A}\delta_{\tau^-}(\mathbf{u}) + \mathbf{b}\|^2,  \nonumber \\
    \text{where } \mathbf{A} =& \begin{bmatrix}
    s_x(\mathbf{u_1}) & s_y(\mathbf{u_1}) \\
    s_x(\mathbf{u_2}) & s_y(\mathbf{u_2}) \\
    \vdots & \vdots \\
    s_x(\mathbf{u_n}) & s_y(\mathbf{u_n})
    \end{bmatrix}, 
    \mathbf{b} = \begin{bmatrix}
    c  \int_{t_0}^{\tau} p(\mathbf{u_1},t) \\
    c  \int_{t_0}^{\tau} p(\mathbf{u_2},t) \\
    \vdots \\
    c  \int_{t_0}^{\tau} p(\mathbf{u_n},t))
    \end{bmatrix}, 
    \label{eq:oject}
    \end{align}
    Finally, the minimum value of \cref{eq:oject} is:
      \renewcommand{\arraystretch}{1}
    \begin{align}
        \delta_{\tau^-}(\mathbf{u}) &=(\mathbf{A}^{T}\mathbf{A})^{-1}\mathbf{A}^{T}\mathbf{b} \nonumber
        \\ & =\frac{1}{\sum s_{x}(\mathbf{u})^2\sum s_{y}(\mathbf{u})^2-(\sum s_{x}(\mathbf{u}) s_{y}(\mathbf{u}))^2}  \nonumber \\ &
        \begin{bmatrix} 
        \sum s_{y}(\mathbf{u})^2\sum SP_x(\mathbf{u}) - \sum s_{x}(\mathbf{u}) s_{y}(\mathbf{u}) \sum SP_y(\mathbf{u}) \\
        -\sum s_{x}(\mathbf{u}) s_{y}(\mathbf{u}) \sum SP_x(\mathbf{u}) + \sum s_{x}(\mathbf{u})^2 \sum SP_y(\mathbf{u})
        \end{bmatrix},
        \label{eq:minimun}
    \end{align}
    where $SP_x(\mathbf{u}) = s_{x}(\mathbf{u})c\int_{t_0}^{\tau} p(\mathbf{u},t) \, dt$ and $SP_y(\mathbf{u}) = s_{y}(\mathbf{u})c\int_{t_0}^{\tau} p(\mathbf{u},t) \, dt$.

\end{document}